\documentclass[conference]{IEEEtran}
\IEEEoverridecommandlockouts
\usepackage[noadjust]{cite}
\usepackage{amsmath,amssymb,amsfonts}
\usepackage{algorithmic}
\usepackage{graphicx}
\usepackage{textcomp}
\usepackage{xcolor}
\usepackage{url} 
\def\BibTeX{{\rm B\kern-.05em{\sc i\kern-.025em b}\kern-.08em
    T\kern-.1667em\lower.7ex\hbox{E}\kern-.125emX}}
\begin{document}

\title{How to predict and optimise with asymmetric error metrics\\
}

\author{
\IEEEauthorblockN{1\textsuperscript{st} Mahdi Abolghasemi}
\IEEEauthorblockA{\textit{School of Mathematics and Physics} \\
\textit{University of Queensland}\\
St Lucia, Australia \\
m.abolghasemi@uq.edu.au}
\and
\IEEEauthorblockN{2\textsuperscript{nd} Richard Bean}
\IEEEauthorblockA{\textit{School of Information Technology and Electrical Engineering} \\
\textit{University of Queensland}\\
St Lucia, Australia \\
r.bean1@uq.edu.au}}

\maketitle

\begin{abstract}
In this paper, we examine the concept of the {\it predict and optimise} problem with specific reference to the $3^{rd}$ Technical Challenge of the IEEE Computational Intelligence Society. In this competition, entrants were asked to forecast building energy use and solar generation at six buildings and six solar installations, and then use their forecast to optimize energy cost while scheduling classes and batteries over a month. We examine the possible effect of underforecasting and overforecasting and asymmetric errors on the optimisation cost. We explore the different nature of loss functions for the prediction and optimisation phase and propose to adjust the final forecasts for a better optimisation cost. We report that while there is a positive correlation between these two, more appropriate loss functions can be used to optimise the costs associated with final decisions. 
%

\end{abstract}

\begin{IEEEkeywords}
prediction, optimisation, solar power, electricity demand, asymmetric errors
\end{IEEEkeywords}

\section{Introduction}
This paper explores the ``predict and optimise'' $3^{rd}$ Technical Challenge which was organised by the IEEE Computational Intelligence Society and Monash University in Australia. The challenge had two main sections as follows: i) forecast the potential power that can be generated for six solar installations and power demand for six buildings for every 15 minutes over the entire month of November 2020, ii) use these predictions as inputs in an optimisation model to schedule charging and discharging batteries as well as classes to minimise the total costs of energy over the same horizon. This type of problem, called {\it predict and optimise}, is becoming increasingly popular in the forecasting, optimisation, and machine learning (ML) literature (\cite{abolghasemi2021effectively, Bengio}). 

One of the important steps for solving the climate change problem is decarbonising energy production by moving towards clean renewable energies and replacing fossil fuel with them. In this regard, wind and solar power are popular as they are relatively cheap to produce and accessible in most places around the world. However, wind and solar power generation are highly dependent on weather conditions that are uncertain. So, we cannot generate them on demand and there are limitations in storing the energy. In order to be able to integrate the renewable energy in the grid and provide a stable network, we need to forecast the electricity demand, electricity price and potential solar power. These forecasts then can be used as inputs into an optimisation model that will determine the optimal time for charging and discharging the batteries, i.e., charge the batteries when price and demand is low, and discharge when electricity is expensive. This will allow us to minimise our consumption from the grid and reduce the costs. 

The ``predict and optimise'' problem is a paradigm in which predictive and prescriptive models need to be used to solve sequential decision making problems where the values are some parameters are unknown and thus should be predicted. This paradigm  is  common in many real world problems. In this problem, although the forecasts in themselves are useful to inform decision makers about the market future, they are not the final goal. The utility of forecasts, e.g., optimising decisions to reduce the electricity costs, is more relevant to decision makers. In fact, the ultimate goal is to reduce the electricity cost using the predictions provided by a model. At first glance, it may seem intuitive that the higher forecast accuracy will lead to a more realistic model and accordingly help the optimisation model to find an optimal solution. If there is a large error in the forecasts, the optimisation model will try to solve an ill-defined model with inputs that are not realistic, thus causing difficulty in finding an optimal solution. However, the association between forecast accuracy and quadratic optimisation cost is neither linear nor symmetric. While the forecasting model aims to minimise the mean absolute error (MAE) or similar linear metrics, the optimisation function aims to minimise a totally different objective with quadratic terms. These two metrics have different objective and it is not always clear how and whether the forecast accuracy is associated with the optimisation costs (\cite{Khabibrakhmanov}). While we know that these two metrics are different, there is little work done to understand the association of forecast accuracy and final optimisation costs. The empirical results are lacking in the renewable energy scheduling problem, and generally in a broader domain of predict+optimise problems. In this paper, we investigate the performance of the forecasting and optimisation models in various scenarios, e.g., overforecasts, underforecasts, and naturally unbiased forecasts,  to shed light on the association of these two seemingly separate problems. Understanding the type of association between these two problems helps decision makers to improve upon whichever model that may be needed most, and more importantly understand how to adjust forecasts or integrate them to get the most desirable solution for the problem at hand.


\section{Problem definition and background}
As indicated in the competition website~(\cite{comp}), the first part of the challenge was concerned with predicting the potential power for six solar panels and electricity demand of six buildings, as accurately as possible. The buildings and solar installations are associated with the Monash University Microgrid, which aims to meet Monash University's electricity needs through solar generation and reduce dependence on the electricity grid.

The second part involved solving a scheduling problem to determine the best decisions and schedule for batteries and student classes in each 15-minute time period. The competition was organised in two phases: phase one which aimed to forecast six buildings demand time series, and six solar power demand for every 15 minute time slot during the entire month of October 2020, i.e., 2976 values forecast for 12 time series and then solving scheduling problems for classes and batteries to determine the optimal decision for charging/discharging batteries and assigning various classes to buildings. The second phase had the same purpose and setting but for November 2020; thus requiring 2880 forecasts for 12 time series and then solving the optimisation models. 

The optimisation part of the competition involved scheduling a set of activities across six buildings for the entire month of October in phase 1 and November in phase 2. These activities are either {\it recurring}, which need to be repeated at the same time and begun within working hours (9 am to 5 pm weekdays) each week, or {\it once-off} that can be scheduled outside of normal working hours sometime during the month. Activities could be scheduled in either small or large rooms as indicated in the example below. Each building has its own corresponding solar, i.e., building 0 has solar 0, building 1 has solar 1, and so on. The microgrid has associated batteries to store the excess energy. These batteries have an efficiency rate and a maximum capacity and can be charged not only with solar panels but also from the grid, where possible. In the competition, the charge and discharge decisions were made for each time period for the maximum capacity; that is, partial charge and discharge decisions were not permitted, in terms of time or power.


Figure~\ref{fig:instance} shows an example provided by the competition organizers of an example instance format. In the first line beginning ``ppoi'' (``predict plus optimize instance'') the five values are the count of buildings, solar installations, batteries, recurring activities, and once-off activities. Each subsequent line beginning ``b'' indexes the buildings (0 to 2) and gives the number of small and large rooms per building. Lines beginning ``s'' map solar installations to buildings, although as the optimization considers only the total net demand, this is not relevant for the optimization task. Lines beginning ``c'' list battery capacity, maximum power, and efficiency. 

Finally, lines beginning ``r'' list the number of large (``L'') or small (``S'') rooms required for the activity, the load required (kW), the duration in fifteen minute periods, the number of ``precedences'' and a list of precedences. Each activity in the list of precedences has to be scheduled within working hours (9am to 5pm) on a weekday before the given activity. For example, activity 0 requires 1 large room, uses 15 kW for 8 periods, and activity 2 must be scheduled on a weekday before activity 0. All recurring activities must be scheduled within working hours on weekdays. Lines beginning ``a'' refer to once-off activities which are omitted from the optimisation in this paper for simplicity.

\begin{figure}
\centering
\begin{verbatim}
ppoi 3 2 1 4 2
b 0 1 2
b 1 1 0
b 2 0 1
s 0 0
s 1 2
c 0 5 2 0.87
r 0 1 L 15 8 1 2
r 1 2 S 8 12 0
r 2 2 L 10 4 0
r 3 1 S 4 4 0
a 0 2 S 8 12 500 100 0
a 1 2 L 8 16 2000 1500 1 0
\end{verbatim}
\caption{Example instance format}
\label{fig:instance}
\end{figure}

Therefore, there is potential power that can be generated from solar panels and a base load that is required for the building operation, resulting in a net load value in each time slot.  The objective is to schedule the activities in the best possible way to minimise the total cost of energy. While it is compulsory to schedule all the recurring activities, the once-off activities can be dropped from the schedule with a certain cost. Since there are smaller number of once-off activities and their impact is relatively small, we have disregarded them in this paper.  The objective function is shown in Equation~(\ref{eq1}). Here $l_t$ refers to the net load in period $t$, $e_t$ is the Victorian wholesale electricity price (which applies in Melbourne where the Monash campus is located), $a_i$ refers to once-off activities, $d_i$ is a boolean decision variable determining whether once-off activity $i$ is included, $o_i$ is a boolean ``out-of-office'' variable, and ${\rm penalty}_i$ is the penalty associated with whether activity $i$ is ``out of office hours''.

\begin{equation}
\begin{aligned}
    O &  = \sum_{t}{\frac{0.25 l_t e_t}{1000}} + 0.005 ({\rm max}_t l_t^2) - \\
    & \sum_{a_i}{(d_i . ({\rm value}_i - p_i {\rm penalty}_i)) }
\end{aligned}
\label{eq1} 
\end{equation}

The objective function contains both a cost of energy (based on the wholesale electricity price) and a  ``peak demand'' charge calculated over the month. (Most retail electricity bills in Australia are issued quarterly.) Unlike other studies of {\it predict and optimise}, the competition energy cost is a quadratic, rather than linear, function. A function with a quadratic term for peak electricity demand may more accurately reflect real-world energy costs. A quadratic term for demand has been used in econometric models of system costs; for example (\cite{Fenrick, Lowry}) performed analysis of hydroelectric system cost using quadratic peak demand terms. In water utilities, energy use is related to the square of the demand as head unit losses are proportional to the square of the flow rate \cite{dasilveira}. Similarly, in electrical energy systems line losses are proportional to the square of the current. Thus, it is logical that cost drivers in energy-related systems can be accurately modelled by a quadratic function and such terms are found to be highly significant in system modelling of cost.

The setting of the above-mentioned {\it predict and optimise} problem is not only useful for renewable energy forecasting and optimising costs but also applicable in other domains with similar settings, e.g., forecasting demand and optimising production lines in manufacturing industries, forecasting workload and scheduling staff. Forecasting is the basis for many managerial decisions. However, it is often not the forecast itself but the utility of the forecast that is of interest for decision makers. The forecasts will be used for making certain decisions that may not be directly associated with the forecast accuracy. While forecast accuracy is measured by an accuracy metric loss function such as MAE \cite{hyndmanfpp3}, these metrics are not informative for decision making in practice and do not directly translate to business decisions. Instead, the added value of the forecasts can be measured through some sort of loss function such as cost or time \cite{abolghasemi2021effectively}.  


The paradigm of predict and optimise is useful when the optimisation problem needs an input that is not fully observed but needs to be predicted. Prediction inevitably has some associated uncertainty with some degree of accuracy. The optimisation which takes inputs from the uncertain forecasts treats them as deterministic values and accordingly determines an optimal solution given the provided inputs. Although several input forecasts can be provided to an optimisation model, there is always a degree of error associated with these inputs and the optimisation models may accordingly change the prescribed solution with respect to the input values. While having reliable forecasts is necessary for finding a reliable optimal solution, it does not guarantee it. 
 
There are studies that have integrated the downstream optimisation in the prediction model and proposed a predictive model with customised loss function that aims to minimise the final cost function (\cite{Bergman2022, demirovic}). For example, Elmachtoub and Grigas (\cite{Elmachtoub}) proposed a smart predict then optimise approach in which they suggested optimising the forecasts with respect to the final optimisation model in which they will be used.  They proposed a new loss function called SPO to train the prediction model. They implemented the proposed model on the shortest path and a portfolio optimisation problem and showed the results are promising. This seems to be a common approach to integrate both prediction and optimisation phases, but in this study we do not aim for this integration because the optimisation function is too complex to be included in the predictive model, and also too computationally expensive for testing several forecast scenarios and obtaining a better optimisation cost. Rather, we are looking to investigate the association between these two and understand how we can adjust the final generated forecasts to optimise the costs associated with a complex objective function. 

The problem of energy system scheduling in stochastic optimization has been studied in Donti et al \cite{donti} who proposed a 2-hidden layer neural network to solve a sample load forecasting and generator scheduling problem. This in turn depended on differentiation in the optimization, using a quadratic programming solver they developed by Amos and Kolter ~\cite{amos}. Later, Cameron et al~\cite{cameron} examined real-world optimization problems in the context of comparing ``two-stage'' (i.e. predict, then optimize) and ``end-to-end'' (i.e. differentiating through the optimization task) approaches. However, the problem investigated by Donti et al was a forecast for only 24 hour ahead generation while the problem under study here is a much more complex mixed-integer quadratic program (MIQP).

There are other research studies that aim to mimic the solutions of optimisation problems with predictive machine learning models. For example, Abolghasemi et al~\cite{abolghasemi2021effectively} proposed to use machine learning models to directly predict the decisions for the first phase of a two-stage stochastic optimisation models. They empirically evaluated their approach on a healthcare supply chain and showed that different loss functions lead to various results that optimise certain costs. These are out of the scope of our current study and we encourage interested readers to see \cite{Bengio} for a good review of the application of machine learning in optimisation.


\section{Data}

\begin{figure}[htbp]
\centerline{\includegraphics[scale=0.15]{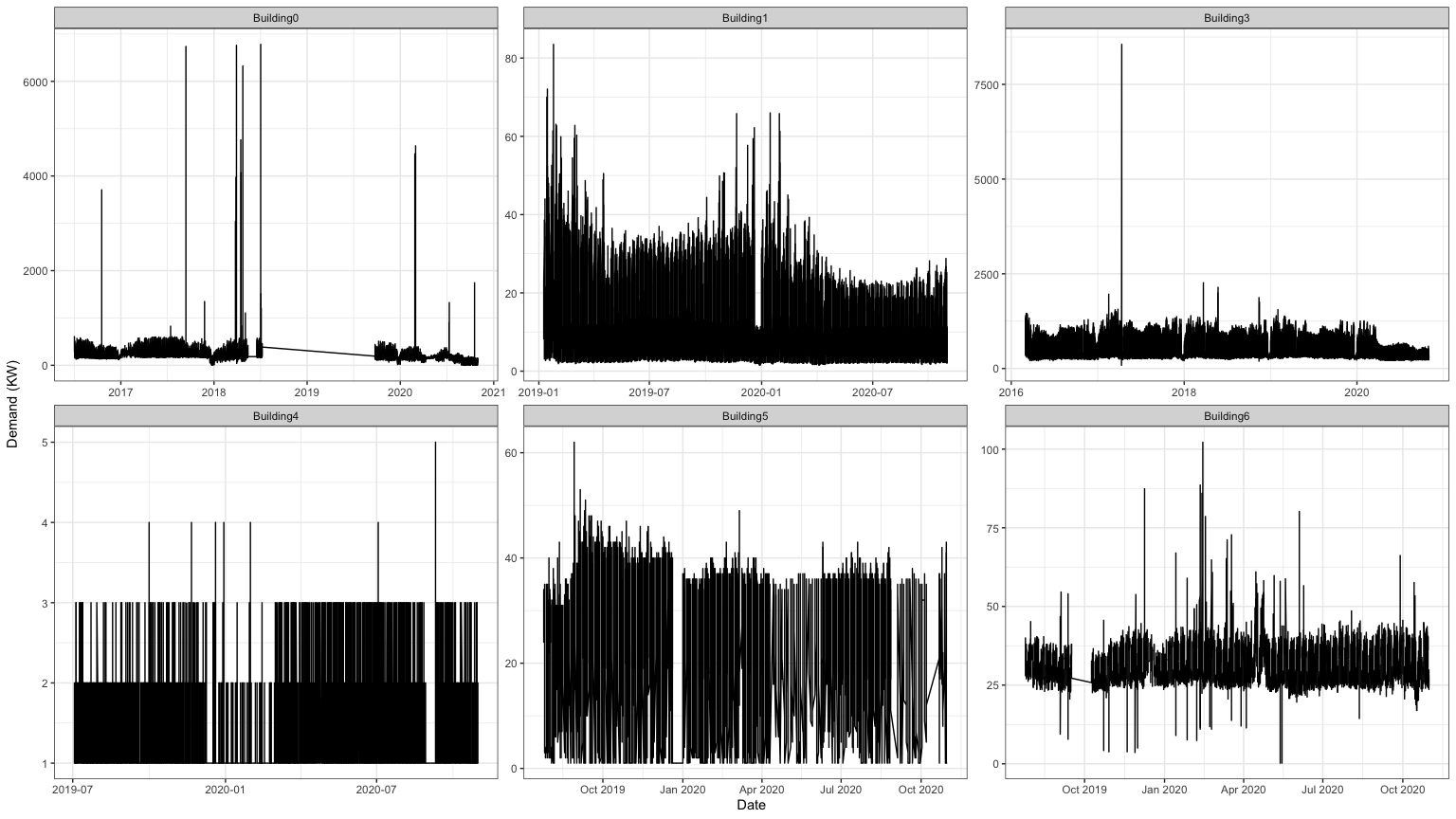}}
\caption{Buildings demand}
\label{fig:buildings}
\end{figure}

\begin{figure}[htbp]
\centerline{\includegraphics[scale=0.15]{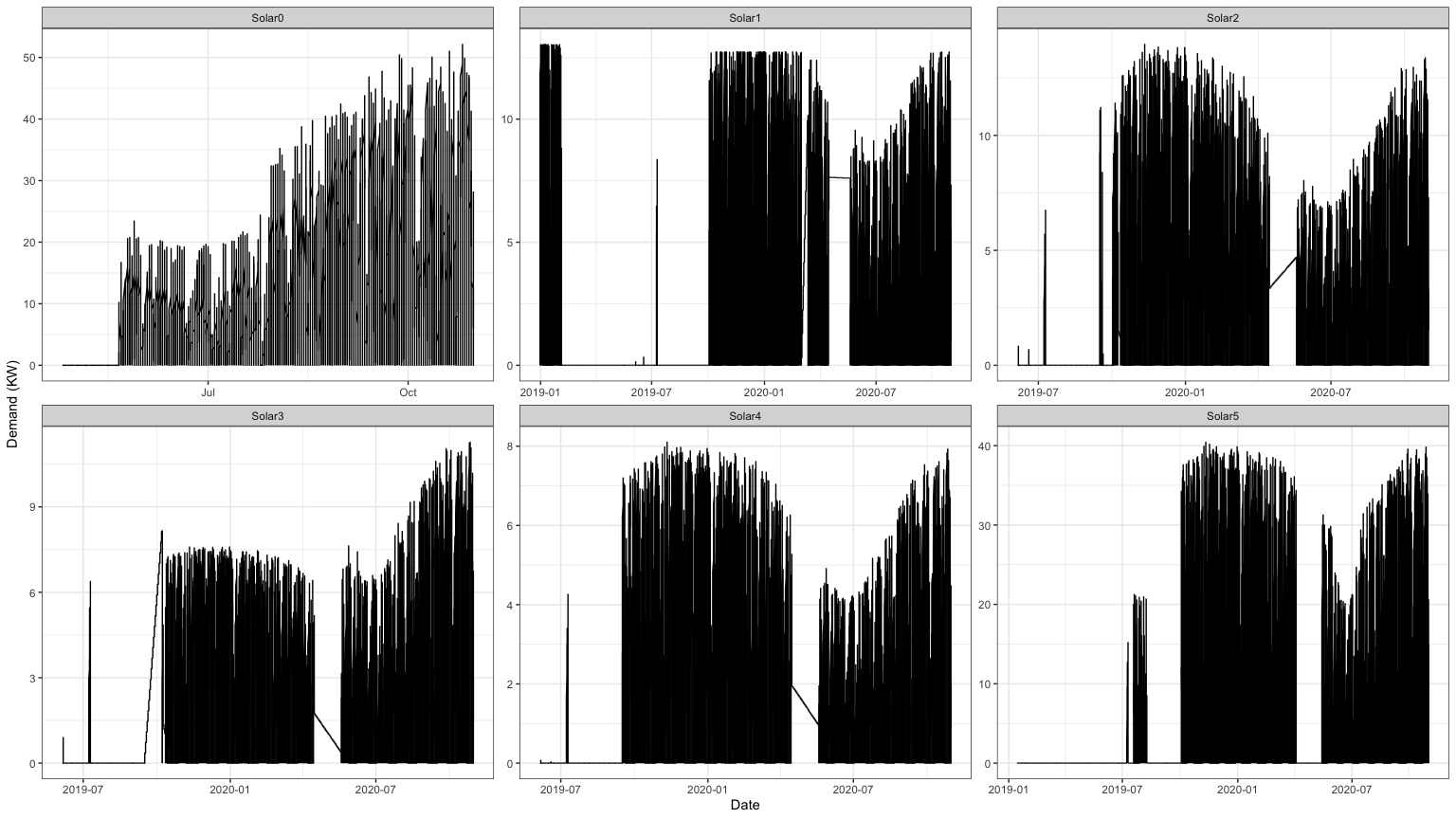}}
\caption{Solar power generation}
\label{fig:solars}
\end{figure}

The datasets consist of time series values for six solar installation outputs and six building demands. The data are recorded at 15 minute granularity with different starting dates. Data were released gradually during the competition. The first phase included data until 30 September 2020 and the second phase included data until the end of 31 October 2020. Therefore, participants needed to provide 2880 period ahead forecasts, for every 15 minutes, for all 12 time series corresponding to November 2020. Figure \ref{fig:buildings} and \ref{fig:solars} show the time series over the entire horizon until the end of 31 October 2020. As we can see in the figures, different buildings have different demand and solar power generation also varies from one series to another. There are many missing or faulty values, around 33\% of the entire data, which are not included in the training data in our experiments. (Thus, when calculating the net load, these values are effectively assumed to be zero values.)

Table~\ref{tab:descdata} shows the descriptive statistics of data after removing the missing values. Building 3 has the largest demand and standard deviations among buildings. Building 4 has the smallest demand, on average. Disparity among the solar panels is smaller in comparison to the building data. The mean of solar power generation varies between 1.12 and 4.39 kW corresponding to solar 5 and solar 0, respectively. While there is no zero demand for buildings, the minimum solar power is zero for all solar panels which occurs before sunrise and after sunset. Solar power generation is highly dependent on the weather conditions and length of the day. Figure \ref{fig:solars} shows the number of periods with zero power for each hour, mapped on each month. As we can see, there are significantly fewer periods without solar power generation during October, November, December, and January where days are longer in the southern hemisphere. 


\begin{table}[htbp]
\caption{Descriptive statistics of data}
\begin{center}
\begin{tabular}{|c|c|c|c|c|c|}
\hline
\cline{2-6} 
\textbf{Series} & \textbf{\textit{Mean}}& \textbf{\textit{Std.Dev}}& \textbf{\textit{IQR}} & \textbf{\textit{Min}} & \textbf{\textit{Max}}\\
\hline
Building 0 & 230.11	&133.54&	108.80&	0.1	&6781.50\\
Building 1& 11.37&	7.91	&8.80&	1.5	&83.50  \\
Building 3& 517.94&	273.81&	415.00	&85.0&	8556.00 \\
Building 4& 1.33	&0.53&	1.00&	1.0&	5.00  \\
Building 5& 24.67&	11.10&	18.00	&1.0&	62.00 \\
Building 6& 30.71&	5.59&	8.40&	0.2&	102.20  \\
Solar 0 & 4.39&	8.86&	4.12&	0.0&	50.41  \\
Solar 1& 1.94&	3.43&	2.41	&0.0&	13.04 \\
Solar 2& 1.88&	3.12&	3.37&	0.0	&13.96  \\
Solar 3& 1.36&	2.21&	1.72&	0.0	&11.04 \\
Solar 4& 3.89&	8.75&	0.97&	0.0	&40.43\\
Solar 5& 1.12&	1.89&	1.55&	0.0&	8.10  \\
\hline
\end{tabular}
\label{tab:descdata}
\end{center}
\end{table}

\begin{figure}[htbp]
       \centerline{\includegraphics[trim=0 100 0 0, clip ,width=7cm]{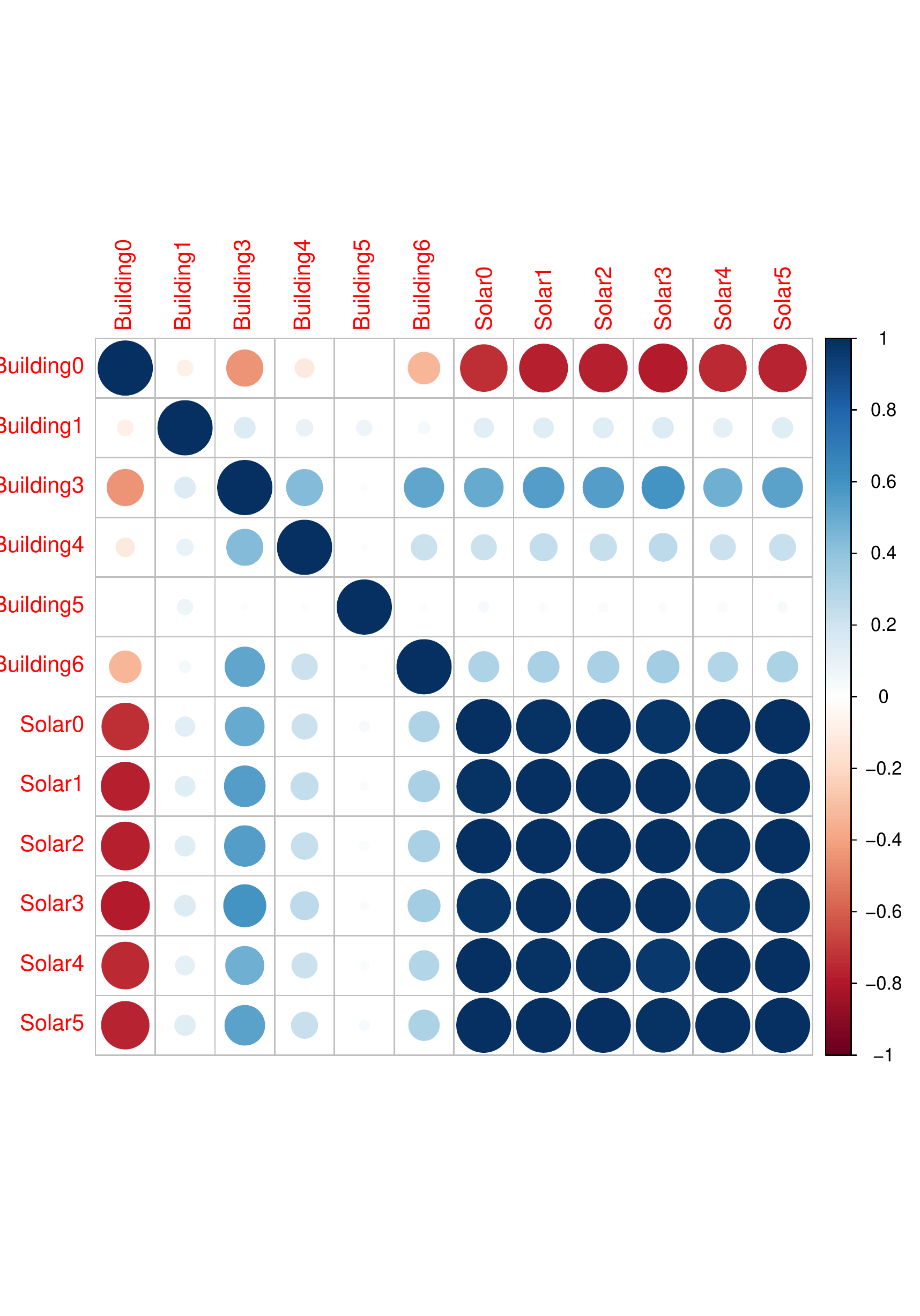}}
\caption{Correlation of 12 time series for October 2020}
\label{fig:corr}
\end{figure}

Figure~\ref{fig:corr} shows a plot of the correlations between the powers for the six buildings and six solar installations (for October 2020 values, where values are not missing). It can be observed that the six solar installations are highly correlated (all correlation coefficients above 0.985) while Building 0 and 3 are also somewhat correlated with the solar generation.

\begin{figure}[htbp]
\centerline{\includegraphics[scale=0.375]{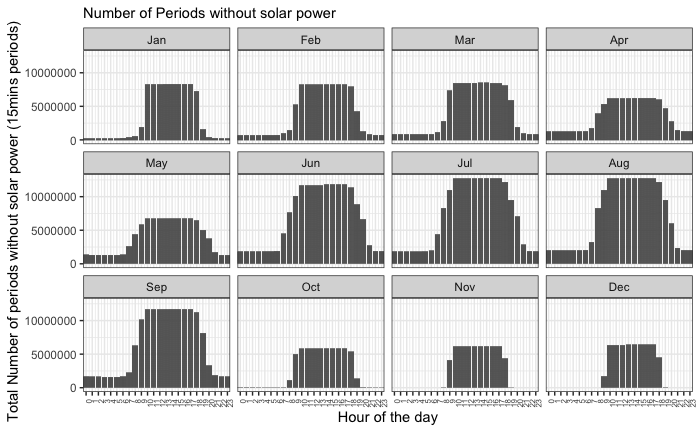}}
\caption{Solar power generation}
\label{fig:solar0}
\end{figure}


The other data provided in the competition was weather data and electricity price data. Weather data was needed to predict the building demand and solar power. Weather data included daily minimum and maximum temperature (C), rainfall (mm) and solar exposure of three weather stations near Melbourne: Moorabbin Airport, Olympic Park, and Oakleigh. This data was available from 1 January 2016. Hourly weather data was also available from the European Centre for Medium-range Weather Forecasting ERA5 dataset via OikoLab.com~\cite{hersbach2018era5}.

Half-hourly wholesale electricity price data was provided for the test set in phase 1 and phase 2. In a real forecasting process, the price would not be available and one needs to forecast it to be able to use it in the optimisation model for scheduling batteries and classes. Scheduling will be a problem that needs to be solved every day to optimise the decision that needs to be taken.  However, this data was made available for the entire months of October and November during the competition, assuming that perfect price forecasting and weather forecasting is available. This is not far from reality, as one day ahead weather forecasting and price forecasting can be executed with a good degree of accuracy. Nonetheless, there will be an additional associated uncertainty and error in the generated forecasts, if the the price and weather data were not available, thus impacting the optimisation solution. 

\section{Methodology}
The aim of this section is to investigate the relationship between forecasting accuracy and optimisation costs for the predict and optimise problem in hand where the forecasts and optimisation models have asymmetric error, i.e., the underforecasts and overforecasts are penalised differently in a complex optimisation model with quadratic cost in objective function. 
Note that the key input to the optimisation model is the \textit{forecasted net base load}, i.e., the total predicted load of buildings minus the predicted solar power for each time slot. That is, we aggregate the predicted buildings demand and deduct it from the aggregated predicted solar power. The result is the net load predicted for the test set, November 2020.  

For the forecasting part, we considered two different types of forecasts.  
In the first type of forecasts, we considered the natural forecasts including the top-performing models in the forecasting phase of the competition, and two newly generated models. 
The first ranked model was a quantile random forest that used weather and calendar features, and the second ranked model was an ensemble of light Gradient Boosting Machine (LGBM) that used calendar and weather features and optimised the parameters.  We also developed two predictive models, LGBM2 and LGBM\_Opt, where we have made some minor changes to the top performing models \cite{bean2022methodology,abolghasemi2021state}. LG2 uses the same setting of the top-performing model but benefits from the ``lightGBM'' package instead of the ``ranger'' package (the second top-performing model used the lightgbm model), while continuing to forecast the 50\% quantile. LGBM\_Opt is the same model where the main hyperparameters of learning rate and number of leaves are tuned. We set the ``num\_leaves'' parameter to 255 and the ``objective'' hyperparameter to ``mae''; that is, attempting to minimise the MAE metric which was the objective of the phase 1 in competition.

The other forecasting models that are ranked among the top seven are also used in our experiment. Each of these methods uses a different model with different input and training setting, thus giving us a range of forecasts generated by a diverse set of architectures to thoroughly investigate the possible association of forecasting and optimisations. We provide a brief review of each model here but interested readers can look at their reports for more details. 
The EVERGi team, ranked third, transformed the data with natural log and used gradient boosting with various features to capture the trend and seasonality  \cite{evergi}. Limmer and Einecke~\cite{honda} who shared the third place with the EVERGi team, used historical and future observations of weather data as well as lags of powers with different orders in a random forest, $k$-nearest neighbours, and gradient boosting. Since we do not have access to their data, we have not considered them in our experiments in this study.  The fifth placed FRESNO team used tree-based models for predicting building energy and ResNET model for predicting the solar power \cite{yuan2021optimal}. The SZU model which ranked sixth, used different models for buildings and solar installaions. For building models they used simple moving average models with different versions, and for solar, ensembles of neural network and other machine learning models~\cite{szu}.
The Stratigakos model, which ranked seventh,  used a quantile regression random forest model with weather data, categorical variables, and preprocessing for some of the noisy data with missing values  \cite{stratigakos2021robust}. The accuracy of the abovementioned models are reported in Table  \ref{tab:accComp}.


For the second type of forecasts, we perturbed the actual values of the building demand and solar to generate scenarios that are either consistently overforecast or underforecast. We generated ten different scenarios ranging from 50 percent underforecast to 50 percent overforecast, with steps of 10\%, comprising 10 scenarios with various accuracy as reported in Table \ref{tab:accPurt}. The reason to generate these scenarios is to have consistent overforecasts and underforecasts that are proportional to the actual values, allowing the optimisation model to look for optimal decisions when it is misinformed. As such, we aim to understand how the optimisation model will react to imperfect biased information, as opposed to natural forecasts where imperfect information are unbiased.

We measure the forecast accuracy using mean absolute scaled error (MASE), and mean absolute error (MAE), as shown in Equations eqref{eq:MASE} and eqref{eq:MAE}, respectively.

\begin{equation}
\text{MASE} = \frac{n-s}{h} \frac{ \sum\nolimits_{t=n+1}^{n+h} {|y_{t}-f_{t}|} } {\sum\nolimits_{t=s+1}^{n} |y_{t}-y_{t-s}|}
\label{eq:MASE}
\end{equation}

\begin{equation}
\text{MAE}= \frac{1}{h} \frac{ \sum\nolimits_{t=1}^{h} {|y_{t}-f_{t}|} } {y_{t}}
\label{eq:MAE}
\end{equation}


where $y_{t}$ is the actual value at time $t$, $f_{t}$ is the predicted value at time $t$, $n$ is the sample size (observations used for training the forecasting model), $s$ is the length of the seasonal period (28 days or 2688 periods), and $h$ is our forecasting horizon. 

Note that MASE was used as the accuracy metric in the competition. MASE is a scale independent metric which makes it suitable for comparing the forecast accuracy when we are measuring the accuracy of many series that have different scales \cite{hyndmanfpp3}. While it made sense to use a scale independent metric such as MASE for measuring the average accuracy of the forecasts in the competition, in this paper it is more sensible to use the MAE metric. This is because here we are dealing only with one time series, i.e., net load, as opposed to the competition where we had 12 time series and a scale independent metric was required to account for differences in magnitude of time series. MAE minimises the median of errors.


Table~\ref{tab:accComp} depicts the accuracy of the top performing models in terms of MAE, MASE, mean overforecast (Mean-O), and underforecast (Mean-U) errors, and the associated total electricity costs. The cost is derived by running the optimisation model described in Bean~\cite{bean2022methodology} which is expanded upon in the next section. The major inputs to the model are the net load forecast and the wholesale pool prices (month-ahead). The Gurobi solver (9.5.0) was used to solve the mixed-integer quadratic program on a system with 11 cores for 168 hours for each scenario.

\begin{table}[htbp]
\caption{Prediction accuracy of the top performing models in the competition}
\begin{center}
\begin{tabular}{|c|c|c|c|c|c|}
\hline
\textbf{}&\multicolumn{5}{|c|}{\textbf{Metrics}} \\
\cline{2-6} 
\textbf{Methods} & \textbf{\textit{MASE}}& \textbf{\textit{MAE}}& \textbf{\textit{Mean-U}}& \textbf{\textit{Mean-O}}& \textbf{\textit{Cost}} \\
\hline
Bean& \textbf{0.64} & 78 & 71& 7 & 34252\\
Abolghasemi &0.74 & 82 & 73& 10& 35003 \\
EVERGi & 0.81 & 90  & 85 & \textbf{4} &	34282\\
FRESNO & 1.00 &102 & 96 & 6 &35464\\
Stratigakos & 0.85 &94 & 88 &6 & 35101 \\
SZU & 0.77 & 68&   {58}& 10 & 35040\\
LGBM2 & 0.88 & 59& \textbf{17}& 42&\textbf{34216}\\
LGBM\_Opt & 0.64 &68 & 57& 10 &34233\\
\hline
\end{tabular}
\label{tab:accComp}
\end{center}
\end{table}

As we can see in ``Table.~\ref{tab:accComp}'', the SZU method is the top performing model among the competition participants in terms of MAE of the net load, even though it ranked sixth in terms of average MASE across all series. (The underlying cause of this is due to an accurate prediction of Building 3 load, which is very large compared to the other time series). This comparison may not be directly applicable to evaluate these models because MASE deals with all series and MAE considers only the single net load. Nevertheless, what matters for the optimisation model is the single net load and its corresponding accuracy both on-average and also across the whole horizon. Therefore, we focus on the MAE metric which is representative of the central tendency of forecasts and its average accuracy. The newly added model, LGBM2, has the lowest MAE among all methods considered in this study. 

Interestingly the lowest optimisation cost corresponds to Bean's model despite the lower performance of the net load forecast, although with a small margin. This indicates that having a more accurate forecast on average does not guarantee a lower cost; rather, the optimisation cost may depend on every single forecast across the entire horizon and the quality of forecast at each time stamp. A large over- or under-forecast may not affect the average of forecast accuracy significantly but it may impact the optimisation cost. Therefore, one needs to generate accurate forecasts for the entire horizon to provide a stable and reliable input to the optimisation model, enabling the optimisation model to perform in its full capacity. The smallest error of overforecast was seen in the EVERGi model followed closely by Abolghasemi and FRESNO models. The smallest underforecast was for SZU model followed by the EVERGi model. 


Table.~\ref{tab:accPurt} depicts the accuracy of various perturbed forecasting scenarios and costs associated with these forecasts. Note that even for the ``Actual'' forecast, the optimality gap is 5.2\% (that is, the lowest possible cost is approximately \$30,800) while for other scenarios the estimated optimality gap ranges from 3.0 to 10.4\% (with the exception of Stratigakos which was solved to optimality).

\begin{table}[htbp]
\caption{Prediction accuracy of the perturbed actual values}
\begin{center}
\begin{tabular}{|c|c|c|c|c|c|}
\hline
\textbf{}&\multicolumn{5}{|c|}{\textbf{Metrics}} \\
\cline{2-6} 
\textbf{Methods} & \textbf{\textit{MASE}}& \textbf{\textit{MAE}}& \textbf{\textit{Mean-U}}& \textbf{\textit{Mean-O}}& \textbf{\textit{Cost}} \\
\hline
Actual & 0 & 0 & 0 & 0& 32494\\
Actual+10 & 0.39 & 56 & 0& 56  &32715 \\
Actual+20 & 0.77 &112 & 0& 112 &33229\\
Actual+30 & 1.16 &167  & 0& 167  &33331  \\
Actual+40 & 1.55 &223 & 0& 223  &33815 \\
Actual+50 & 1.93 & 279& 0 & 279 &36338  \\ 
Actual-10 & 0.39 & 56 & 56& 0 &32514 \\
Actual-20 & 0.77 &112 & 112& 0 &32646 \\
Actual-30 & 1.16 &167 & 1670& 0 &32920 \\
Actual-40 & 1.55 &223 & 223& 0 & 33842\\
Actual-50 & 1.93 & 279& 279 & 0 &34204 \\
\hline
\end{tabular}
\label{tab:accPurt}
\end{center}
\end{table}

The ``Actual'' column is the actual values of building demands and solar powers after realisation. We also generated 10 different scenarios where we generate five scenarios of overforecasts and underforecasts by adding and subtracting  10\%, 20\%, 30\%, 40\%, and 50\% to the actual values, respectively.

The results of Table.~\ref{tab:accPurt} are different and show another angle of the problem. Although the average accuracy of various models for over forecast and underforecasts are symmetric, the associated optimisation costs are not. 

%
%

\begin{figure}[htbp]
\centerline{\includegraphics[width=7.5cm]{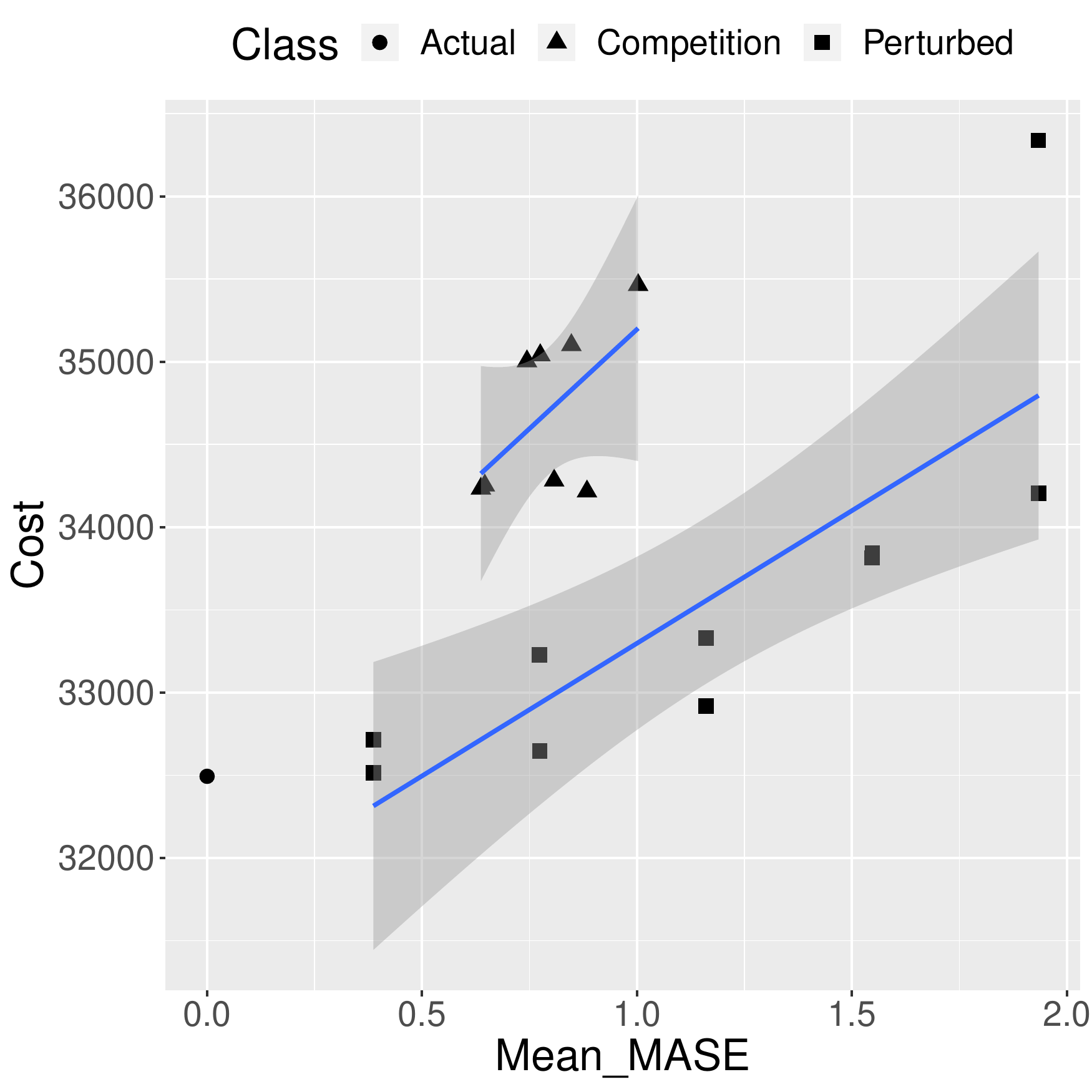}}


\caption{Relationship between forecast accuracy and energy cost}
\label{fig:acc}
\end{figure}

Fig.~\ref{fig:acc} shows the association between predictive accuracy of net load for all the investigated scenarios (expressed in terms of the competition metric, MASE over the twelve time series) and their corresponding optimisation costs. The two lines give a linear regression between cost and mean MASE for the two classes of Competition and Perturbed forecasts.  

We assessed the correlation coefficient between the final cost and several metrics across (1) the perturbed actual net load, plus the actual net load (11 forecasts), (2) the competition forecasts, plus the actual net load (9 forecasts). The resulting correlation coefficients are displayed in Table~\ref{tab:corr_metric_cost}.


\begin{table}[htbp]
\caption{Correlation of error metrics with optimisation cost}
\begin{center}
\begin{tabular}{|c|c|c|}
\hline
\textbf{}&\multicolumn{2}{|c|}{\textbf{Methods}} \\
\cline{2-3} 
\textbf{Metrics} &
\textbf{\textit{Perturbed}}& \textbf{\textit{Competition}} \\ 
\hline
MASE & 0.815 & 0.897  \\
MAE  & 0.815 & 0.901 \\
Mean-U  & 0.051 & 0.803  \\
Mean-O  & 0.684 & 0.083 \\ 
Mean Residuals  & 0.355 & -0.648  \\ 
Std Dev Residuals  & 0.815 & 0.909  \\ 
Skewness Residuals  & 0.308 & -0.255  \\ 
Kurtosis Residuals  & na & 0.044  \\ 
\hline
\end{tabular}
\label{tab:corr_metric_cost}
\end{center}
\end{table}

For perturbed forecasts, the highest correlation was observed for the standard error metrics, MASE of the time series and MAE of the net load. However, this perturbation is not reflective of the kind of errors that would be seen in an actual forecasting situation.

For the competition forecasts we found that similar correlations were observed for ``Mean-U'' or the magnitude of the mean underforecast, MASE and MAE. The correlation for ``Mean-U'' is much stronger than the correlation for ``Mean-O'' or the mean overforecast. This is in line with our impressions regarding the relative importance of underforecasting in the competition; a large underforecast in a peak hour period may lead to a large jump in the peak load over the month, which is quite costly.

\begin{figure}
    \centering
    \includegraphics[width=7.5cm]{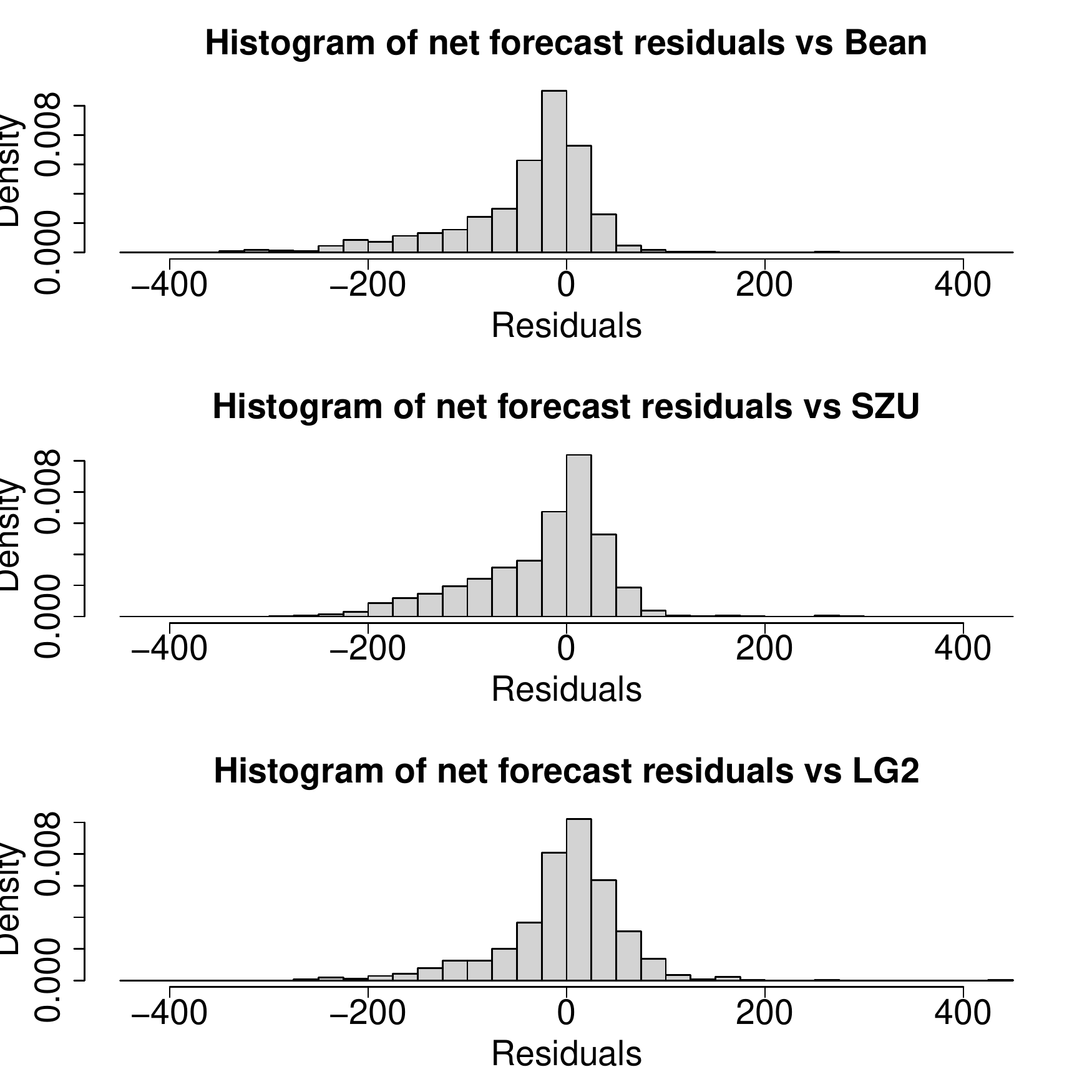}
    \caption{Histograms of net forecast residuals versus three competition forecasts}
    \label{fig:hist}
\end{figure}

Fig. \ref{fig:hist} shows a histogram of the residual errors for three forecasts (Bean, SZU and LG2) for the net load forecast. The moments of the residuals (mean, standard deviation, skewness and kurtosis) are shown for each plot. The Building 3 residuals have a strong effect on the net forecast residuals, as the scale of the Building 3 time series is so large compared to the other 11 time series. In particular, the LG2 forecast for Building 3 is quite symmetric relative to the other forecasts.

Hence, the skewness of the net forecast with the LG2 method is very low compared to other methods at -0.39; that is, closer to a Gaussian distribution. The skewness of the residuals of the other competition entries ranges from -0.76 to -1.12. We discuss the effect of the skewness in Section~\ref{disc}.

We now describe the optimisation method used in more detail and examine possible improvements to the metric.

\subsection{Optimisation} 
For the optimisation assessment, we disregarded the one-off activities for simplicity. This is more computationally feasible, allowing us to run multiple scenarios. It is also realistic as there are not many once-off activities. The original competition required competitors to optimise 10 scenarios, five large and five small. Each of the these 10 scenarios contained two batteries. In our assessment as we are introducing multiple forecasts, we focused on only the first of the scenarios, named ``small0''. This scenario had 50 recurring activities and 20 once-off activities. In the formulation we chose, this meant that each of the 50 recurring activities had to be assigned to one of 160 15-minute periods in the first week of the month, then each was allocated between one and three small or large rooms. 


The objective function of the optimisation part as shown in \ref{eq1} minimises the total energy cost that includes a quadratic form, the square of the maximum load. As noted this accurately reflects cost drivers in many real-world scenarios, although we are unaware of any tariffs which use this particular formulation or are cost-reflective in this manner.



We describe in general terms five different optimisation approaches ``conservative'', ``forced discharge'', ``no forced discharge'', ``liberal'' and ``very liberal''. For each approach, the input is a single, unweighted forecast which is the result of our forecasting procedure. 

\begin{itemize}
    \item The {\it Conservative} approach effectively ignores the forecast. It assigns the recurring activities to periods such that over the first week (and thus, the first month), the peak load is minimized. If the electricity price is relatively flat, this effectively minimizes energy cost while effectively ignoring the underlying net load. This was the starting point for the optimization approach of Bean. If there is no {\it a priori} knowledge of the underlying net load or electricity prices, this may be the correct approach.

    \item {\it Forced discharge} forbids any charging in peak hours, and forces at least one of the two batteries present to be discharging in every peak period (that is, 9am to 5pm on weekdays). This approach attempts to avoid spikes in the objective function caused by an underforecast of the net load. A large underforecast of net load is very costly as the optimizer, or scheduler, will consider it ``safe'' to schedule many activities in that period, which will drive up the peak load and increase the quadratic term in the cost function.

    \item {\it No forced discharge} forbids any charging in peak hours, but the optimizer decides whether to discharge or do nothing in those hours.

    \item {\it Liberal} allows charging in peak, but an extra constaint is added: the maximum of recurring load (classes) plus charge effect for each period cannot exceed the maximum of the recurring load over all periods. Again, the intent is to avoid problems caused by underforecasting. When the solver input for a particular period has a large underforecast and a low electricity price, the solver may choose to schedule battery charging for that period. The effect is again to drive up the peak load and increase the overall cost while the optimizer believes the opposite is occurring.

    \item {\it Very liberal} allows charging over peak and does not attempt to control the maximum of recurring plus charge effect. If the forecast was perfect, this would be the correct approach.
\end{itemize}

In the optimization, firstly the ``conservative'' approach was used to generate a range of possible ``warm starts'' for the solver. For example, 46 different starting solutions were generated for the ``small0'' case studied here, using the ``conservative'' approach which attempts to schedule all and minimize the peak load. Then the solver is given the forecast, chooses the best starting point from among all the warm starts, and attempts to solve minimizing the objective function of Equation (\ref{eq1}) which contains the quadratic cost term.

An ad-hoc analysis of each of the five approaches showed that ``no forced discharge'' performed the best over the ``Phase 1'' scenarios. The choice of which of the five approaches to use could only be based on the experience of the competitors in ``Phase 1'' or October 2020. For the analysis in the paper, we used the ``no forced discharge'' approach.

Note that the order of the five approaches are presented in reflecting our relative confidence in the forecasts. If we have no confidence in our forecast and any forecast is as likely as any other, ``conservative'' is most appropriate while for perfect forecasts ``very liberal'' is most appropriate. Also, instead of perturbing the forecast to mitigate the effects of overforecasting, we simply assessed the approaches across the known values of October 2020 to mitigate the effects of underforecasting. In the following section we examine a more systematic approach, albeit one requiring knowledge of how a forecast generally performs.

\section{Discussion}\label{disc} 
Khabibrakhmanov et al~\cite{Khabibrakhmanov} provides a discussion of the effects of underforecasting and overforecasting in solar forecasting. They consider how a provided forecast can be corrected to minimize a given cost function. 

The previous sections have shown that there is no direct relationship between the usual metrics of MAE and the value of the cost function (both calculated with respect to the actual net load). Using measures of under or over forecasting provides a closer correlation but it is clear there are other factors at work.


The concepts studied in Khabibrakhmanov et al provide a way to determine a similar concept for this study. Instead of a solar forecast alone, we examine how a forecast of net load in the microgrid (building demand minus solar generation) is related to the cost function.

Equation~\ref{equ} measures the mean absolute error (divided by two) (Khabibrakhmanov et al Equation 6). This is equivalent to measuring the MAE of the net load.

\begin{equation}
\begin{aligned}
    U & = \frac{1}{2N} \sum_{i=0}^{N-1}{|y_i-p_i|}
\end{aligned}
\label{equ}
\end{equation}

If instead of measuring the absolute error of the net load forecast, we add first and third order perturbation terms as in Equation~\ref{eqv} (Khabibrakhamnov et al Equation 7) we can study the effect of under or over forecasting. Here $y_i$ refers to actual net load and $p_i$ refers to predicted net load. Khabibrakhmanov et al explain that minimizing the cost function $V$ leads to determining the bias and slope for an optimal linear correction for $V$, $p_i = \beta + \alpha x_i$.

\begin{equation}
\begin{aligned}
    V  = & \frac{1}{2N} \sum_{i=0}^{N-1}{(y_i-p_i)^2}+\frac{1}{N}\gamma\sum_{i=0}^{N-1}{(y_i-p_i)}+ \\
    & \frac{1}{3N}\epsilon\sum_{i=0}^{N-1}{(y_i-p_i^2)^3}
    \end{aligned}
    \label{eqv}
\end{equation}

As a result of the competition, we have a number of energy forecasts derived through at least two processes (random forest, LightGBM) by independent methods by at least six different teams. We solved for $\gamma$ and $\epsilon$ in Equation~\ref{eqv} across the eight forecasts plus the actual net load, in Table~\ref{tab:accComp}, by maximizing the Pearson correlation between the cost function $V$ and the costs found in Table \ref{tab:accPurt}, and also solved across the perturbed forecasts. The correlation coefficients found were 0.955 (for the perturbed cases) and 0.881 (for the competition cases), indicating that there is a strong correlation between the forecast accuracy and optimisation costs. This confirms that the predictive and prescriptive models operate hand in hand and in order to minimise the optimisation cost, one would need to have an accurate forecast regardless of the optimisation method used. This association is strong for the typical predict and optimise problem that we have investigated. 

\begin{figure}
    \centering
    \includegraphics[scale=.5]{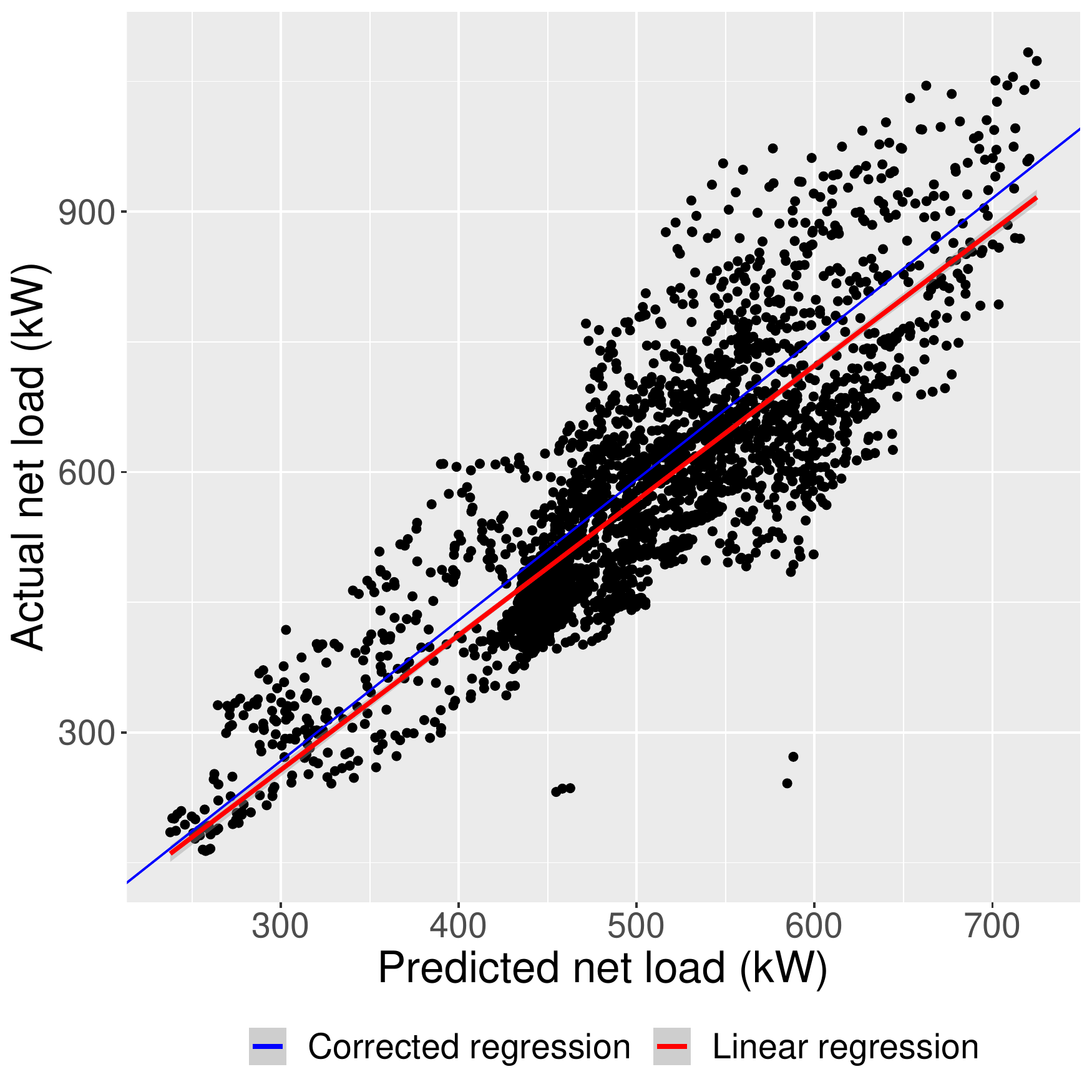} 
    \caption{Linear correction for predicted net load versus actual net load to minimize cost. The red line shows the original linear regression of predicted versus actual net load, which minimizes the sum of squares between actual and predicted net load of Equation \ref{equ}. The blue line shows the corrected regression line which minmizes the asymmetric error cost function of Equation \ref{eqv}.}
    \label{fig:netload}
\end{figure}

Normalizing the forecast and actual values to have mean 0 and standard deviation 1, for the competition cases, we obtained $\gamma = 1.37$ and $\epsilon = 0.58$. We plot the unnormalized values of forecast and actual net load in Figure~\ref{fig:netload}. 

As both of these estimated parameters $\gamma$ and $\epsilon$ have positive sign, this indicates a forecast error where the actual net load exceeds the forecast net load, or an ``underforecast'', incurs an extra cost using the optimization method we have chosen (that is, ``no forced discharge''). 
The third order perturbation term in Equation \ref{eqv} is related to the skewness of the residuals. As previously noted, the skewness of the competition prediction residuals is also (negatively) correlated with the cost; all these skewness values are negative as the predictions underestimated the actual net load.


In order to further minimize the ultimate cost, we could correct the forecast, or predicted values, using a linear cost correction as seen in the blue line in Figure~\ref{fig:netload}, or find an ``end-to-end'' method for solving the task which may be computationally much more complex. We may also experiment with different optimization parameters or attempt to linearize the quadratic term in the objective function, as many competitors attempted to do. Further work is necessary across different scenarios to determine the best cost correction to apply.

\section{Conclusion}

We examined solutions from the $3^{rd}$ IEEE Technical Challenge on predict and optimise which was about energy forecasting and scheduling activities. We analysed the correlation between forecast accuracy, overforecasting, and underforecasting and examined the associated optimization costs to understand how forecast accuracy impacts the optimisation decision. We then discussed how one can adjust the forecast for minimizing a quadratic cost function. 

After assessing five approaches for optimization against a month where actual net load was known, we chose an approach named ``no forced discharge'' for a particular battery operation strategy in the optimization. This approach was assesssed on multiple different forecasts, both from the competition and perturbations of the actual load. We found that the traditional error rate metrics of mean absolute error (MAE), as well as measures of under and overforecasting, provide some explanation for the connection between the metrics and ultimate cost in case when the objective function is quadratic and forecasts errors are asymmetric. Examining the effects of asymmetric cost functions in solar forecasting, we derived a linear correction to apply to the prediction to attempt to minimize final costs across sets of competition entries and perturbed entries.


Econometric assessments have demonstrated that the cost drivers in some energy systems can be modelled well by quadratic cost functions, and quadratic terms can be justified by physical processes.  In the context of the IEEE competition, it appeared that underforecasting load had a significant asymmetric effect on the final cost function, and several methods were suggested to correct for this.

Further research is required to determine improved methods for solving the scheduling problem as an ``end-to-end task'' or alternatively developing a robust method to bias the forecast which is an important input to the optimization. We may also assess different optimization methods of other competition entrants e.g. ~\cite{abolghasemi2021state}, ~\cite{honda} and \cite{evergi} as the solutions found during this research have not been shown to be optimal in any sense. One interesting future research avenue is to build a machine learning model that can map the inputs of the forecasting model in the prediction part directly to the outputs of the optimising models. A multi-input multi-output machine learning model can be used to evaluate the efficacy of this approach. 

\section*{Acknowledgments}
We are thankful for the participants of the third technical challenge of the IEEE-CIS competition for making their codes and reports publicly available at https://ieee-dataport.org/competitions/ieee-cis-technical-challenge-predictoptimize-renewable-energy-scheduling. 

\end{document}